\documentclass[a4paper,conference]{IEEEtran}
\ifCLASSINFOpdf
  \usepackage[pdftex]{graphicx}
\else
\fi
\usepackage{amsmath}
\usepackage{comment}

\begin{document}
%

  
\title{%
On the use of learning-based forecasting methods for ameliorating fashion business processes\\
\LARGE A position paper}

\author{\IEEEauthorblockN{Geri Skenderi}
\IEEEauthorblockA{Department of Computer Science\\
University of Verona\\
Verona, Italy}
\and
\IEEEauthorblockN{Christian Joppi}
\IEEEauthorblockA{Humatics Srl\\
Verona, Italy}
\and
\IEEEauthorblockN{Matteo Denitto}
\IEEEauthorblockA{Humatics Srl\\
Verona, Italy}
\and
\IEEEauthorblockN{Marco Cristani}
\IEEEauthorblockA{Department of Computer Science\\
University of Verona, Humatics Srl\\
Verona, Italy}
}


%


\maketitle

\begin{abstract}
The fashion industry is one of the most active and competitive markets in the world, manufacturing millions of products and reaching large audiences every year. A plethora of business processes are involved in this large-scale industry, but due to the generally short life-cycle of clothing items, supply-chain management and retailing strategies are crucial for good market performance. Correctly understanding the wants and needs of clients, managing logistic issues and marketing the correct products are high-level problems with a lot of uncertainty associated to them given the number of influencing factors, but most importantly due to the unpredictability often associated with the future. It is therefore straightforward that forecasting methods, which generate predictions of the future, are indispensable in order to ameliorate all the various business processes that deal with the true purpose and meaning of fashion: having a lot of people wear a particular product or style, rendering these items, people and consequently brands fashionable. In this paper, we provide an overview of three concrete forecasting tasks that any fashion company can apply in order to improve their industrial and market impact. We underline advances and issues in all three tasks and argue about their importance and the impact they can have at an industrial level. Finally, we highlight issues and directions of future work, reflecting on how learning-based forecasting methods can further aid the fashion industry.
\end{abstract}


%
\IEEEpeerreviewmaketitle

\section{Introduction}
The technological evolution of the fashion industry has been nothing short of remarkable. Electronic retail, also known as e-commerce, is such a normal commodity nowadays that it's hard to believe that the first e-commerce websites were introduced less than thirty years ago. E-commerce is evergrowing and it is predicted to be worth 24.5\% of the whole apparel industry by 2025 \cite{ecom_statista}. Constant innovation has rendered the Fashion \& Apparel industry one of the most lucrative industries in the world \cite{fashion_statista}, and it is expected to keep rising in the following years. Nevertheless, this industry has been widely criticized for its large waste generation \cite{pollution2022}, because of phenomena like overproduction and product returns~\cite{lane2019uses}. The main reason for this can be tied back to customer dissatisfaction~\cite{masyhuri2022key}, whether that relates to size, color, style, or textile quality. As a result, for the sector to successfully regulate environmentally friendly production methods and be as efficient as possible, it must become fully customer-centric and understand its clients at a profound level. Exploiting the extensive advances in Machine Learning solutions in the past few years is one of the better ways to do this. 

In this paper, we explore the particular utility of forecasting, a discipline that has remained closely related to retail ever since its early days throughout the 20th century \cite{yule1927,box2015time} until recent years, where multiple forecasting systems are present \cite{taylor2018prophet,salinas2020deepar,Oreshkin2020N-BEATS:}. Specifically, we will detail how forecasting models that are based on machine learning can massively improve business processes in fashion, from supply chain and inventory management to marketing policies. The main reason why we argue in favor of these models is that they can bypass several limitations of established and well-known time series analysis methods like AR, MA or ARIMA \cite{box2015time}. To discuss these limitations and discuss why solutions to them are crucial, we will consider three different tasks in this paper: 
\begin{enumerate}
    \item New product demand forecasting, where the future sales of a new product have to be estimated without having any historical data;
    \item Short-term new product forecasting, where the future sales of a recently published product have to be estimated based on only a few observations; 
    \item Product recommendation, where the future preferences of a user have to be predicted based on their past purchases or similarity with other users.
\end{enumerate}

In the following chapters, we will first talk about the advances in each respective task and then argue, using concrete results on different datasets, why these tasks are important and why they can accurately and efficiently be solved using deep learning methods. We then talk about open issues in each task and wrap up the paper by discussing future possibilities of forecasting with deep learning for fashion. We hope that this paper can motivate further research in this field by laying out a series of arguments backed up by concrete results, while also acting as a short survey on these particular tasks that can be considered "edge-cases" of forecasting in fashion.  

\section{New product demand forecasting}
Sales forecasting is one of the most typical and important forecasting applications in fashion retail~\cite{intelligentFashionBook,KashiSalesSurvey}. The ability to predict client demands and behavior can make a huge difference in commercial activity, especially when vast volumes of items need to be controlled, for a variety of economic and financial reasons. While the forecasting of time series with a known historical past is an extensively studied topic~\cite{box2015time,hyndman_athanasopoulos_2021,benitez_salesForecastingReview_2021}, another practical and much harder case has received very little attention: the forecasting of \emph{new items} that the market has yet to witness. Experts create judgmental projections in these circumstances~\cite{hyndman_athanasopoulos_2021}, taking into account the characteristics of the freshly designed product as well as information on what is currently trending in the market to make an educated guess. In simpler words, the \emph{demand} for a new product is foretasted based on an expert's experience.

Under this last formulation, it starts to become clear why machine learning can help in this scenario: It is possible to design models in such a way that they learn this judgmental process. The intuition behind models that can help in solving this task is that generally, new products will sell comparably to similar, older products; consequently, these models should be able to understand similarities among new and older products and understand how trends evolve with time. We will now demonstrate how this task has been tackled in the literature, in order to then provide insight on what can be additionally be done. 

In~\cite{singh_fashion_2019}, a variety of machine learning models are taken into consideration such as boosting algorithms (XGBoost, Gradient Boosted Trees) and Neural Networks (MLP, LSTM). The authors propose to simulate the expert judgment by utilizing the textual attributes related to category and color, and merchandising factors such as discounts or promotions. In this case, it is possible to train a model using a batch of products, where of course the object's release date is also clearly stated, such that the models can process each ground truth time series and learn inter-product correlations based on their common textual attributes. This is something that classical methods like ARIMA \emph{simply cannot achieve}, as they cannot be fit on multiple series in different time frames, but most importantly they \emph{cannot be utilized} when we have no past observations. In this scenario, relying on a machine learning solution is the most efficient and performing solution.

A big problem with the work of~\cite{singh_fashion_2019}, is that they do not consider any visual information. Even though textual attributes are important, they can never create a good enough representation for a fashion item like visual attributes. A picture is, after all, worth one thousand words. In~\cite{ekambaram2020attention}, the authors make a big step forward and use an autoregressive RNN model that takes past sales, auxiliary signals like the release date and discounts, textual embeddings of product attributes, and the product image as input. The model then uses an additive attention mechanism~\cite{bahdanau2016neural} to understand which of the modalities is the most important to the sales and combines all these attended features into a feature vector which is fed to a GRU~\cite{cho2014learning} decoder. This Recurrent Neural Network based architecture is indeed an accurate model for expert judgments. Each product is represented by its visual, textual, and temporal attributes, which creates fine-grained groups of similar products. The autoregressive component of the RNN allows the model to take past predictions into account, therefore generating a forecast which is reliant on the initial guess. One of the most important aspects of this work is that the attention mechanism also provides an interpretability component, demonstrating which modality is the most important at each forecasting step. The authors try out different variations of the architecture and their best performing model, Cross-Attention Explainable RNN, reports less than 40\% weighted Mean Absolute Percentage Error (wMAPE) in some partitions of an \emph{unshared, proprietary dataset}.

A few issues that were not considered in the aforementioned papers, are the concept of trends and popularity, as well as a benchmark where the new product demand forecasting task could be tested. In~\cite{skenderi2021well}, the authors tackle all of these problems, proposing a Transformer-based, multimodal architecture that can also process exogenous Google Trends signals. These signals bring the idea of the fashion expert judgment full circle, as now the model can reason based on the past performances of similar products in the dataset, as well as the online popularity of the different attributes that make up the product. Additionally, to avoid the error accumulation problem of autoregressive RNNs, the authors propose a generative approach where the whole demand series is forecasted in one step. Finally, the authors render their dataset, named Visuelle, public. This is an important aspect as it promotes further research on the methodological aspect of this challenging task, by relying on a dataset that is actually based on the real sales of a fast-fashion company. On Visuelle, the GTM-Transformer model of~\cite{skenderi2021well} reports a wMAPE of 55,2\%, compared to 59\% of Cross-Attention Explainable RNN, as shown also in Table~\ref{tab:demand}. In~\cite{joppi2022pop} the authors introduce a data-centric pipeline able to generate a novel exogenous observation data, the POtential Performance (POP) signal, comparing directly the new product image with the images uploaded on the web in the past. The POP signal fed into the GTM-Transformer model represents the state-of-the-art on this task, with a wMAPE of 52,39\%. In~\cite{papadopoulos2022multimodal} a novel Multimodal Quasi-AutoRegressive deep learning architecture (MuQAR) for the New Product Demand Forecasting task has been proposed. The model uses attributes and captions automatically extracted from the product image by the use of image classification and captioning methods. Attributes are used both to broaden the set of google trends and as input to the model together with the image captions. MuQAR model reports a wMAPE of 52,63\% (Table~\ref{tab:demand}).

\begin{table}[t]
\small
\centering
\caption{New Product Demand Forecasting results on Visuelle in terms of Weighted Mean Absolute Percentage Error (WAPE) and Mean Absolute Error (MAE). The lower the better for both metrics.}
\begin{tabular}{|l|l||c|c|}\hline
    \textbf{Method} & \textbf{Input } &\textbf{WAPE} & \textbf{MAE} \\ \hline \hline
\emph{Attribute KNN}~\cite{ekambaram2020attention} 2020&[T]&59,8&32,7\\ \hline
\emph{Image KNN}~\cite{ekambaram2020attention} 2020&[I]&62,2&34,0\\ \hline
\emph{Attr+Image KNN}~\cite{ekambaram2020attention} 2020&[T+I]&61,3&33,5\\ \hline
\emph{GBoosting}~\cite{friedman_2001} 2001&[T+I+G]&63,5&34,7\\ \hline
\emph{Cat-MM-RNN}~\cite{ekambaram2020attention} 2020&[T+I+G]&65,9&35,8\\ \hline
\emph{CrossAttnRNN}~\cite{ekambaram2020attention} 2020&[T+I+G]&59,0&32,1\\ \hline\hline
MuQAR~\cite{papadopoulos2022multimodal}&[T+I+G+A+C]&52,6&28,8\\ \hline\hline
\textbf{GTM-Transformer}~\cite{skenderi2021well}&[T+I+G]&55,2&30,2\\ \hline
\textbf{GTM-Transformer}~\cite{joppi2022pop}&[T+I+POP]&\textbf{52,4}&\textbf{28,6}\\ \hline

\end{tabular}

\label{tab:demand}
\end{table}

\begin{figure*}
    \centering
    \includegraphics[width=\textwidth]{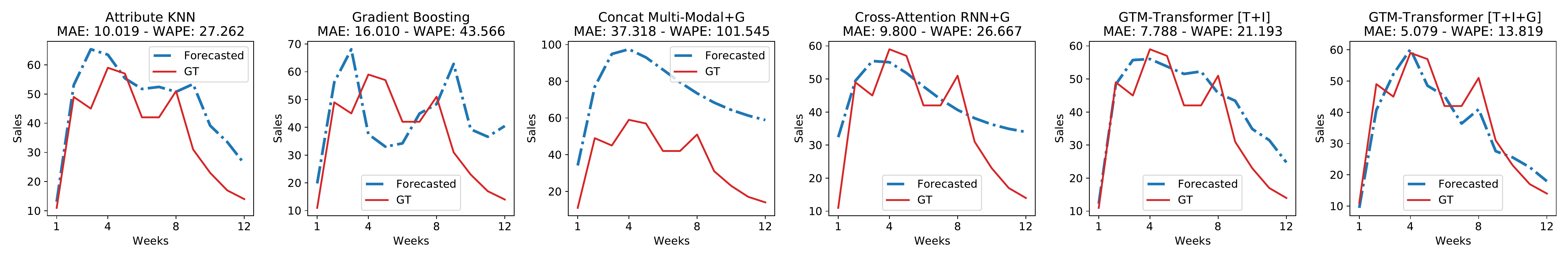}
    \includegraphics[width=\textwidth]{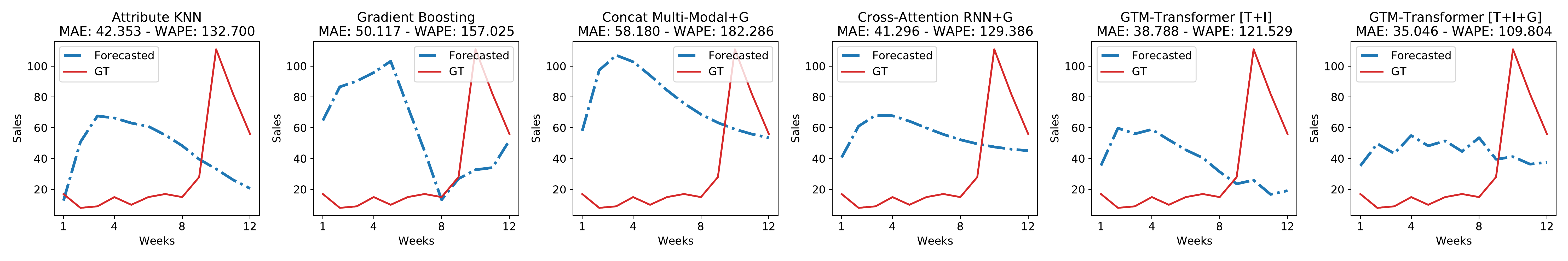}
    \caption{Qualitative results for the New Product Demand Forecasting of two different products on VISUELLE.}
    \label{fig:qlt_visuelle} 
\end{figure*}

\begin{figure*}
    \centering
    \includegraphics[width=0.49\textwidth]{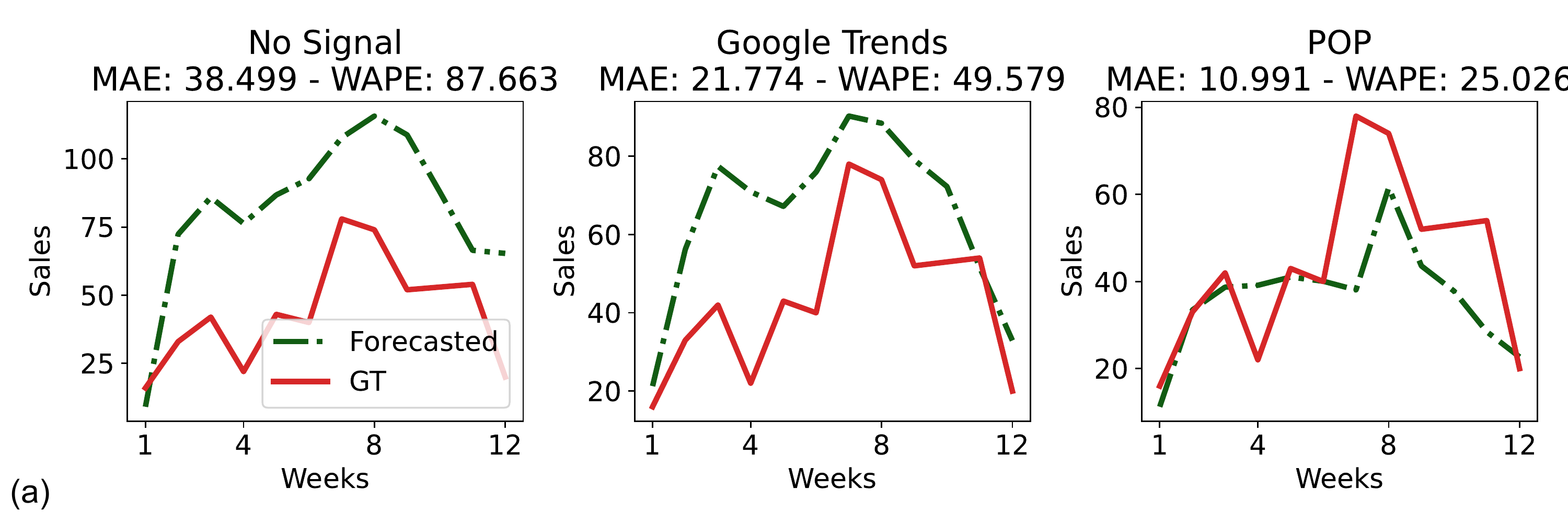}
    \includegraphics[width=0.49\textwidth]{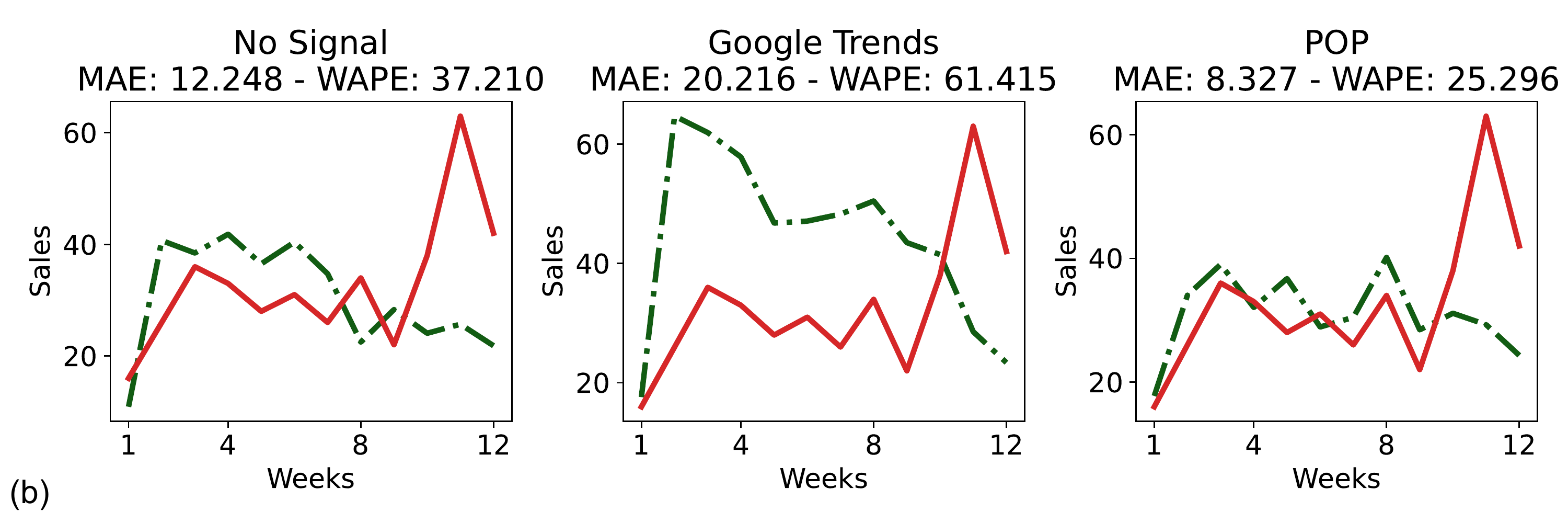}
    \caption{Qualitative results for the New Product Demand Forecasting of two different products on VISUELLE with different exogenous signals.}
    \label{fig:qlt_visuelle_exsignals} 
\end{figure*}

These results demonstrate that the problem of new product demand forecasting benefits greatly from machine learning and computer vision. It is also very important to consider multi-modality and augment the model reasoning power by using exogenous data that can act as a proxy for the notion of popularity. Fig. \ref{fig:qlt_visuelle} shows different demand forecasting systems at work when they perform the forecasting of new product sales. The best-performing architecture is able to produce forecasts that closely resemble the testing ground truth series, showing how multi-modality and exogenous information helps in improving forecasting performance. Fig. \ref{fig:qlt_visuelle_exsignals} shows the results with different exogenous signals used with the GTM-Transformer architecture. POP provides the most accurate curve. To provide a concrete example as to why performing well on this task is important, we report here a result from~\cite{skenderi2021well}. A typical supply chain operation related to new product demand forecasting is the first order problem, i.e., ordering a number of products that matches the sum of future sales, without exceeding or underestimating in the best case. By using a typical supply chain policy instead of forecasting based first order, the company would lose 683,372\$, while the seemingly small improvement of 5\% in wMAPE terms, results in more than 156,942\$ spared. It is clear to see how this forecasting task can massively help in sorting out initial doubts regarding early supply chain operations, which can translate into a lot of money-saving and efficient decisions.

\section{Short-observation new product forecasting}
Short-observation new product forecasting (SO-fore) aims at predicting future sales in the short term, having a past statistic given by the early sales of a given product. In practice, after a few weeks from the delivery on market, one has to sense how well a clothing item has performed and forecast its behavior in the coming weeks. This is crucial to improve \emph{restocking policies}~\cite{maass2014improving}: a clothing item with a rapid selling rate should be restocked to avoid stockouts. The same reasoning can be applied to the opposite case, where a product with a slow-selling rate might not necessarily be a focus point for restocking in the near future. Restocking policies are typically processes that are always associated with high uncertainty, because of constantly shifting market dynamics. While it is simpler, in principle, for companies to make an initial guess when having to perform their first restock, i.e., new product demand forecasting, managing restocks while a product is selling can get complicated in the short term, until the end of the product's lifetime.

To assess both of these problems, we argue that it is useful to consider two particular cases of the SO-fore problem, based on the estimated lifetime of the product under consideration. Consider a product $p$ that will sell for a pre-defined amount of timesteps $t$ (these can be days, weeks, months, etc.). Given only a few observations, $t': t' \ll t$, the first set-up would be SO-fore$_{t'-({t-t'})}$, in which the observed window is $t'$ weeks long and the forecasting horizon is $t-t'$ weeks long (the remaining, unseen observations), required when a company wants to implement few restocks~\cite{fisher2001optimizing}. This corresponds to forecasting the sales based on the few available observations until the end of the product's lifetime, focusing on the estimation of major restocks.

The second set-up would be SO-fore$_{t'-1}$, where the forecasting horizon changes to a single timestep (at a time), and is instead required when a company wants to take decisions on a frequent basis, as in the \emph{ultra-fast fashion} supply chain~\cite{taplin2014global,choi2014fast}. This corresponds to forecasting the sales for each timestep based on the $t'$ ones before it, focusing on the estimation of a minor, more frequent, restocks.

In these scenarios, because we have past observations, we can consider \emph{some} classical forecasting approaches, such as Naive method~\cite{FPAP2} (using the last observed ground truth value as the forecast) or Simple Exponential Smoothing (SES)~\cite{hyndman2008smoothing, FPAP2}. It is still impossible, in practice, to fit models that make assumptions about the process that generates the time series such as AR, MA, or ARIMA models. Even though in theory having more than one observation is sufficient, to estimate statistically significant parameters a larger sample ($t'=50$ \cite{box2015time}) is required. Therefore, without relying on ML approaches, the forecasting practitioner in this case would have to rely on methods that calculate simple statistics based on the last observed values. On the other hand, similarly to the task of New product demand forecasting, learning-based approaches can exploit inter-product similarity and also the similarity between the initial sales period in order to provide much more accurate forecasts. This problem is of course, simpler than new product demand forecasting, since knowing the initial $t'$ observations can greatly benefit forecasting, without acting purely in a judgmental guessing framework such as the models from \cite{skenderi2021well}.

In~\cite{skenderi2022multi}, the authors introduce Visuelle 2.0, a second version of the homonym dataset, where the time series data are disaggregated at the shop level, and include the sales, inventory stock, max-normalized prices, and discounts. This permits to perform SO-fore considering every single store. The authors then implement on Visuelle 2.0 one ML and one DL approach from~\cite{ekambaram2020attention}:k-Nearest Neighbors and Cross-Attention Explainable RNN. Given that each product in the dataset is known to sell for $t=12$ weeks, it is assumed that two weeks of sales are observed ($t'=2$ weeks). In both cases, i.e., SO-fore$_{2-1}$ and SO-fore$_{2-10}$ the authors show that the RNN approach outperforms the others by a significant margin, reaching a wMAPE of respectively 23,20 and 32,25. The classical forecasting approaches on the other hand give much poorer performances. It is also worth noticing that exploiting image embeddings proves to be crucial to further improve the performances in the case of longer horizons. For details, we refer to the work in ~\cite{skenderi2022multi}.

Short-observation new product forecasting is a task of high importance especially in the industry of fast-fashion, for several reasons. Firstly, it is necessary to have estimates about restocking policies when the product lifecycle is so dynamic. Secondly, given that most fast-fashion companies have multiple shops, it is necessary to understand how different products sell in each shop. Finally, utilising forecasting approaches continuously is very important for marketing reasons in specific shops, because if the company knows how much a given product will sell in advance, they can try to make it sell more or less by changing the item's position within the shop. It is therefore very important to have benchmarks that can capture these different elements, in order to advance this particular area of fashion forecasting. We believe that Visuelle 2.0, as mentioned before, contains a lot of data and information and it is the first of its kind, but it is important for industrial practitioners to try and create other datasets that might present different forecasting challenges in the short term.

\section{Product recommendation}
So far we have explored problems that deal a lot with the supply chain. Of course, this is not the only aspect that a company would have to manage in the real world, given that the main goal is to sell a particular product. The biggest issue with fashion is that it represents a form of expressiveness and like any other form of expressiveness, it is quite subjective and often defined by different societal biases. It is therefore useful to continuously have an idea as to what particular clients like and tailor new products that are specific to some client base. This is by no means a new idea, yet given the abundance of choices when looking at clothing items, a good recommendation system is required to correctly sort, order, and deliver relevant product content or information to users. Effective fashion recommender systems may improve the purchasing experiences of billions of customers while also increasing supplier sales and income. Deploying such systems in the fashion industry is not trivial, often because the purchase history for a customer is very sparse and because most customers do not define themselves based on a single style. The biggest issue of these mining and retrieval systems is that often they might categorize users as actually belonging to a particular style or group of clients and immerse them in a "bubble of information", where the user keeps getting similar recommendations. On the other hand, the systems must also maintain a relatively conservatory notion of novelty and not recommend random things to a user.

We believe that several of these issues can be bypassed if the temporal aspect were to be considered instead of treating product recommendation as a static problem. Most recommender approaches deployed in practice are instances of collaborative or content filtering, where the systems often relied on user or item similarity to perform recommendations. As an illustrative example, consider a client who typically buys sportswear, but recently bought a suit jacket and a shirt. Considering user-user similarity, the system will probably provide a recommendation that is related to sportswear, since that is what other similar users have provided. The same thing can be said about the items, where most of the population is made up of a particular type of article. What is often missing in these cases, is the notion of the time of purchase of each product. Using temporal continuity in recommender systems is nothing novel, as seen in the field of research of session recommendations \cite{li2017neural,ren2019repeatnet}, but actually relying on the temporal properties of purchases to characterize users is something that is not deployed in typical systems. We like to refer to this notion as similarity-over-time.

\begin{figure}
    \centering
    \includegraphics[width=\columnwidth]{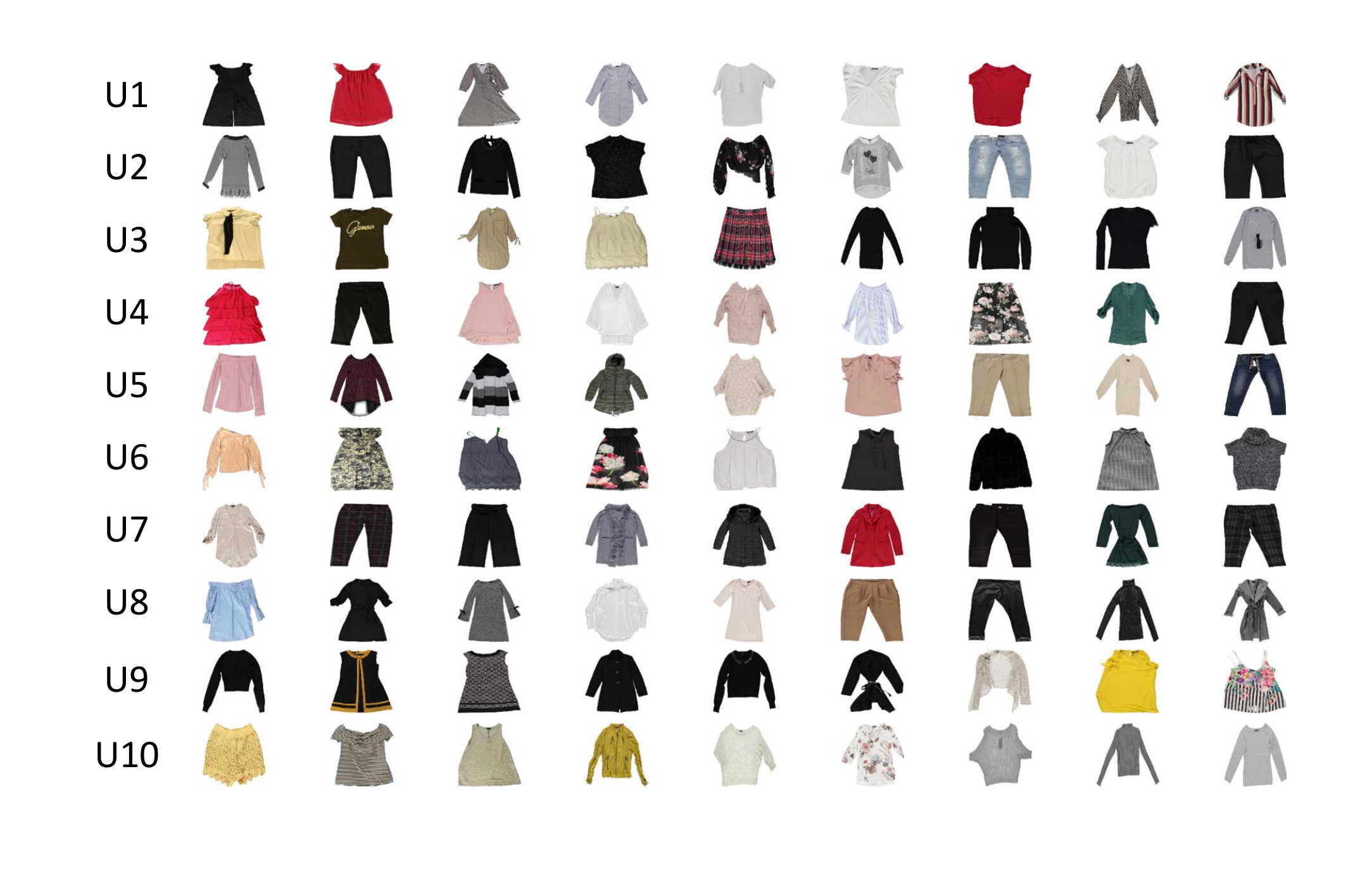}
    \caption{A random sampling of users purchases. Personal styles do emerge: users 1 and 10 have no trousers in their logs, user 6 has bought almost short sleeves and no trousers, while user 7 seems to prefer long sleeves and several trousers; user 10 has a marked preference for light yellow-grayish colors.}
    \label{fig:example_cust1}
\end{figure}

If the recommendation task was cast as a forecasting task, a model would also have to keep in mind the temporal relationship between the purchases (or ratings) and similarity-over-time. In this scenario, a simple idea could be to give more weight to later observations or, given enough data, understand the seasonal trends of the users. It is, therefore, crucial to have data on which to learn or try these approaches. A real-case scenario of such datasets for fashion is Visuelle 2.0 \cite{skenderi2022multi} and the Amazon reviews dataset~\cite{ni2019justifying}. A graphical example of customer data from Visuelle 2.0 is reported in Fig.~\ref{fig:example_cust1}, where it is visible that some users have marked preferences: The recent purchases of user 10 are all grey-ish items.

Product recommendation as suggested here would consist in defining a particular time index $t_{\text{rec}}$, when the historical data of all the past purchases (older than $t_{\text{rec}}$) of all the customers will be taken into account. Therefore, two types of inferences will be possible: 1) to suggest which product (or category, or attribute) $z_{k}$ a specific customer $u_{i}$ could be interested in; a positive match will be in the case of an effective purchase of $z_{k}$ (or some item which is in the category $z_{k}$ or that expresses the attribute $z_{k}$) by $u_{i}$ after time $t_{\text{rec}}$; 2) same as before, but including a specific time interval $T_{\text{buy}}$ within which the customer will buy. In practice, a positive match will be in the case of an effective purchase of $z_{k}$ (or some item which is in the category $z_{k}$ or that expresses the attribute $z_{k}$) by $u_{i}$ \emph{in the time interval} $]t_{\text{rec}},t_{\text{rec}}+T_{\text{buy}}]$. This creates a system that can provide more novelty and also interaction with users, since as the purchase data grow, so does the system's ability to understand the user's preferences over time. Because this is the idea of casting recommendation as forecasting ~\cite{hu2015web,fayyaz2020recommendation} is inherently quite multi-modal, the use of machine learning techniques, especially deep learning, can provide optimal solutions, even though in principle the recommendation itself can still be carried out by slightly tuning standard techniques. This interplay could be certainly explored using the Visuelle 2.0 dataset.

\section{Conclusion}
\subsection{Future work}
In general, learning forecasting models for fashion products is a very challenging task. The sales dynamics often have high variance and they are quite different from one company to another. A primary focus should be put into creating datasets that contain sales information from multiple platforms, in order to create forecasting models that generalize better. Of course, a model has to be fit to a specific series to get the best performance, but for small start-ups, it is very important to have analytical support as early as possible. This means having models that can be fine-tuned on small datasets and so far this issue is yet to be explored in industrial forecasting scenarios.

New product demand forecasting can be further improved by reasoning on other shared, prior factors of the products. The authors in \cite{skenderi2021well} suggest popularity, but item availability and also price series play an important role. Considering the price, future work should pay attention to multivariate forecasting of the sales and the price. This allows companies to understand the dynamics of the sales and the evolution of the price, which aids discount campaigns and profit understanding. The same can be said about Short observation of new product forecasting.

The temporal recommendation is a relatively new field, but one promising direction of future work is the forecasting of irregular and sparse time-series. Since most users buy items in a sporadic way, it is hard for any model to learn temporal dependencies other than an order for most users. These models should therefore reason also in terms of similarity-over-time, recommending items that similar users with more purchases bought at that time. In this way, the recommendation can improve interactions with users, but also avoid recommending only evergreen products and present new products to their users.
\subsection{Ethical issues}
From the point of view of societal impact, we believe that forecasting approaches can be highly beneficial for reducing pollution since fashion is the third most polluting industry in the world. Having a precise estimation of sales or popularity can improve the situation by solving supply chain issues. The economical aspect is also very important since accurate forecasts can lead to millions of spared USD, as reported in \cite{skenderi2021well}. Nevertheless, forecasting models do bring about ethical implications. The first one is the reduced attention toward industry experts because if the models start to perform well, companies might start cutting departments that only deal with judgmental forecasts. In the case of product recommendations, information bubbles can be produced and limit both the user and the company, since a lot of products will not be seen or recommended. Finally, basing all decisions based on forecasts leads to a phenomenon that is well known in finance, which is that they become self-confirmed, i.e, the forecasts become real because every action was based on them. For this reason, it is preferred to use these tools as decision-making helpers rather than oracles and use probabilistic forecasting to understand the uncertainty related to the predictions.
\subsection{Final remarks}
In this position paper, we discuss about learning-based forecasting methods for the fashion industry, and we argue why their use is absolutely necessary for different scenarios. Machine learning is increasingly becoming more important in other aspects of FashionAI such as clothing generation, virtual try-on, and product search, yet supply-chain operations and business processes related to restocking and selling are often still tackled with specific company policies. We believe that forecasting solutions should be carefully implemented at different stages of the product lifecycle and guide companies towards more informed decision-making, ultimately leading to a more efficient and productive world of fashion.

\subsection{Acknowledgements}
This work has been partially supported by the Italian Ministry of Education, University and Research (MIUR) with the grant “Dipartimenti di Eccellenza” 2018-2022.






%


\bibliographystyle{IEEEtran}
\bibliography{main}

\end{document}